\documentclass[runningheads]{llncs}

\usepackage{eccv}

\usepackage{eccvabbrv}

\usepackage{graphicx}
\usepackage{booktabs}

\usepackage[accsupp]{axessibility}  

\usepackage{svg}
\usepackage{multirow}

\usepackage[pagebackref]{hyperref}

\usepackage{orcidlink}

\begin{document}

\title{ARTA: Adaptive Mixed-Resolution Token Allocation for Efficient Dense Feature Extraction}


\author{David Hagerman$\dagger$, Roman Naeem$\dagger$, Erik Brorsson, Fredrik Kahl \& Lennart Svensson\\
\texttt{\{david.hagerman, nroman, erik.brorsson, lennart.svensson, fredrik.kahl\}@chalmers.se} \\
}

\authorrunning{D.~Hagerman et al.}

\institute{Chalmers University of TechnologyChalmers University of technology, 412 96 Gothenburg, Sweden
\email{kontakt@chalmers.se}\\
\url{http://www.chalmers.se/}}

\maketitle

\begin{abstract}
We present \textbf{ARTA}, a mixed-resolution coarse-to-fine vision transformer for efficient dense feature extraction. Unlike models that begin with dense high-resolution (fine) tokens, ARTA starts with low-resolution (coarse) tokens and uses a lightweight allocator to predict which regions require more fine tokens. The allocator iteratively predicts a semantic (class) boundary score and allocates additional tokens to patches above a low threshold, concentrating token density near boundaries while maintaining high sensitivity to weak boundary evidence. This targeted allocation encourages tokens to represent a single semantic class rather than a mixture of classes. Mixed-resolution attention enables interaction between coarse and fine tokens, focusing computation on semantically complex areas while avoiding redundant processing in homogeneous regions. Experiments demonstrate that ARTA achieves state-of-the-art results on ADE20K and COCO-Stuff with substantially fewer FLOPs, and delivers competitive performance on Cityscapes at markedly lower compute. For example, ARTA-Base attains \textbf{54.6 mIoU} on ADE20K in the $\sim$100M-parameter class while using fewer FLOPs and less memory than comparable backbones.
\keywords{Segmentation \and Adaptive token allocation \and Mixed-resolution}
\end{abstract}

\footnote{$\dagger$ These authors contributed equally to this work.}

\section{Introduction}

In semantic segmentation tasks, not all pixels are equally important. In a typical scene, large regions are often homogeneous (e.g., sky, road, or walls) and exhibit little spatial or semantic variation, making dense per-pixel computation unnecessary. In contrast, regions with high object density or class boundaries require higher spatial resolution for accurate predictions. This uneven distribution of semantic content motivates architectures that adaptively allocate higher spatial resolution to more informative regions of the image.

Most computer vision architectures process images using a uniform grid of square patches, assigning equal computational cost to each patch regardless of its semantic content. For example, in Vision Transformers \cite{dosovitskiy_image_2020}, it is common to divide the input image into fixed-size patches (e.g., 16×16), each of which is embedded into a token, applying the same representation capacity to both simple and complex patches. Convolutional neural networks such as ConvNext \cite{woo_convnext_2023} similarly apply filters over regularly spaced patches, implicitly treating all areas of the image with equal importance. While some recent models \cite{ziwen_autofocusformer_2023, lu_content-aware_2023, tang_dynamic_2023} introduce mechanisms for content-aware processing or adaptive computation, the majority of widely used architectures still rely on uniform spatial partitioning, which does not align with the uneven distribution of information in real-world images.

While uniform spatial partitioning is inefficient, most non-uniform approaches still follow a high-to-low resolution processing pipeline. In these architectures, the input image is initially represented at high resolution, and information is gradually compressed through pooling, striding, or token merging. As a result, all image patches, regardless of their semantic importance, are initially processed at the finest spatial granularity. This leads to unnecessary computation in homogeneous or uninformative patches and limits the potential efficiency gains of token pruning or merging in later stages. An ideal approach would avoid assigning high-resolution capacity to patches that do not require it in the first place, allocating computational resources only where semantic complexity demands it.

Beyond where computation is spent, there is also the question of how representational capacity is used within each token. When a token aggregates pixels from multiple semantic classes, its feature vector must encode the class mixture (which classes are present and in what proportions), the spatial layout and extent of each class within the patch, and the patch’s overall position. By contrast, if a token predominantly corresponds to a single class, its representation can devote more dimensions to modeling intra-class variation and higher-order semantics rather than resolving class boundaries. Constraining tokens to be largely class-specific improves the efficiency of the representation, allowing models with modest width to rival the semantic expressiveness of wider baselines.

To address these computational and representational inefficiencies, we propose \textbf{ARTA} (\textbf{A}daptive Mixed-\textbf{R}esolution \textbf{T}oken \textbf{A}llocation), a two-stage encoder. In Stage 1, a lightweight allocator adaptively assigns token density across the image through three hierarchical allocation rounds. Starting from coarse tokens, it predicts a semantic class-boundary score for each token and allocates additional finer-grained tokens to patches whose scores exceed a low threshold, increasing resolution only where boundary evidence is present. In subsequent rounds, scoring is re-applied only to the newly allocated finest-resolution tokens, reducing allocator compute by avoiding repeated dense evaluation over the full token set. Repeating this process yields a mixed-resolution token set that concentrates spatial detail near class boundaries and avoids unnecessary allocation in uniform regions, ensuring that no patch is processed at higher resolution than needed. In Stage 2, we refine the resulting mixed-resolution tokens with a deeper hierarchical encoder to capture higher-level semantics.
To summarize, our main contributions are:
\begin{itemize}
    \item A lightweight, class-boundary scoring allocator for adaptive mixed resolution token allocation. Starting from coarse tokens, it iteratively re-scores and allocates finer-grained tokens to patches containing class-boundaries, increasing token density in semantically rich regions while leaving uniform areas at coarse granularity.
    \item \textbf{ARTA:} A two-stage encoder that couples the proposed mixed-resolution token allocation with a deeper hierarchical refinement network. Mixed resolution attention enables interaction between coarse and fine tokens, concentrating computation on semantically complex patches.
    \item We validate ARTA on ADE20K, COCO-Stuff, and Cityscapes, achieving state-of-the-art results on ADE20K and COCO-Stuff with substantially fewer FLOPs, and competitive performance on Cityscapes at lower compute.
\end{itemize}

\section{Related Work}

\subsection{Token dropping and merging}

Many methods improve transformer efficiency by removing tokens deemed uninformative or by aggregating nearby/similar tokens \cite{rao_dynamicvit_2021,pan_scalable_2021,bolya_token_2022,marin_token_2023,long_beyond_2023,wei_joint_2023,kim_token_2024,chang_making_2023}. These fine-to-coarse strategies typically start from a dense high-resolution grid of tokens and progressively reduce the number of tokens by pruning or merging similar tokens to gain efficiency. This yields a content-adaptive token density, but only through token reduction. The spatial resolution never exceeds that of the initial grid, and training still processes an initially dense token set. In contrast, ARTA follows a coarse-to-fine strategy that starts from coarse tokens and allocates additional tokens only where finer detail is needed, increasing resolution on demand rather than uniformly.

\subsection{Adaptive downsampling}

A complementary direction makes tokenization or routing content-aware, for example via saliency-driven multi-resolution grids, attention-guided sampling, or per-input policy decisions \cite{ronen_vision_2023,fayyaz_adaptive_2022,meng_adavit_2022}. For dense prediction, AutoFocusFormer (AFF) \cite{ziwen_autofocusformer_2023} proposes point-based local attention with balanced clustering and a learnable neighborhood-merging module to support segmentation heads. Other works merge/share or prune tokens based on semantics or difficulty \cite{lu_content-aware_2023,tang_dynamic_2023}. While these approaches introduce adaptivity, most still start from a dense tokenization and then select, share, or prune tokens thereafter. As a result, they generally do not increase spatial token density on demand, and training compute remains dominated by the initial high-resolution token set.

\subsection{Learned upsampling operators}

Another line of work studies learned upsampling operators that incorporate local context. Dynamic upsampling operators such as DySample \cite{liu_learning_2023} and frequency-aware fusion modules such as FreqFusion \cite{chen_frequency-aware_2024} operate on dense feature maps. DySample learns sampling locations for each output position, while FreqFusion applies adaptive low-/high-pass filtering and offsets to boundary sharpness during upsampling. These approaches focus on how to upsample by improving the upsampling operator itself, and they typically upsample densely across spatial locations. In contrast, ARTA addresses what to upsample by starting from a coarse tokenization and allocating additional tokens only when required, producing a mixed-resolution representation that keeps homogeneous patches compact. Therefore, learned upsampling operators are orthogonal to ARTA.

\subsection{Coarse-to-fine architectures}

Several coarse-to-fine architectures follow an overview-to-detail schedule, redistributing capacity toward coarse scales \cite{zhu_parameter-inverted_2024,lou_overlock_2025}. OverLoCK \cite{lou_overlock_2025} begins with a coarse global context derived from low-resolution features, and later refines the representation using fine-grained attention. These architectures alter how capacity is distributed across scales but generally keep resolution schedules fixed and input-agnostic rather than content-driven within an image.

Beyond encoder design, boundary-refinement post-processing methods such as SegFix \cite{yuan_segfix_2020} improve boundary quality by redirecting boundary pixels toward more reliable interior predictions using learned offsets. In contrast, ARTA incorporates boundary awareness directly into the encoder through class-boundary scoring and hierarchical token allocation. SegFix operates as a post-processing step on model outputs, whereas ARTA controls where spatial resolution is allocated during feature extraction.

Overall, most prior work either reduces token count after a dense beginning, improves upsampling or fusion on dense feature maps, or redistributes capacity with fixed multi-scale plans. A gap remains for architectures that start from coarse tokens and adaptively allocate higher token density only where semantic complexity warrants it, while maintaining mixed-resolution representations that interact across scales during dense feature extraction.

\section{Method}

We present \textbf{ARTA}, a two-stage hierarchical encoder that constructs mixed-resolution feature representations by adaptively allocating token density to semantically complex image patches. ARTA is designed to encourage tokens to correspond to a single semantic class, rather than mixing multiple classes within the same token. This is achieved by identifying tokens likely to overlap semantic (class) boundaries and selectively allocating additional tokens to only those patches, while keeping homogeneous patches compact.

ARTA differs from token-pruning or token-merging approaches that begin with a dense high-resolution grid and later reduce token count. Instead, ARTA starts from coarse tokens and increases spatial resolution only where needed. This avoids ever allocating high-resolution tokens to semantically simple patches and enables a principled and efficient use of computation.

\subsection{Overview}

\begin{figure}[t]
  \begin{center}
  \includegraphics[width=0.98\textwidth]{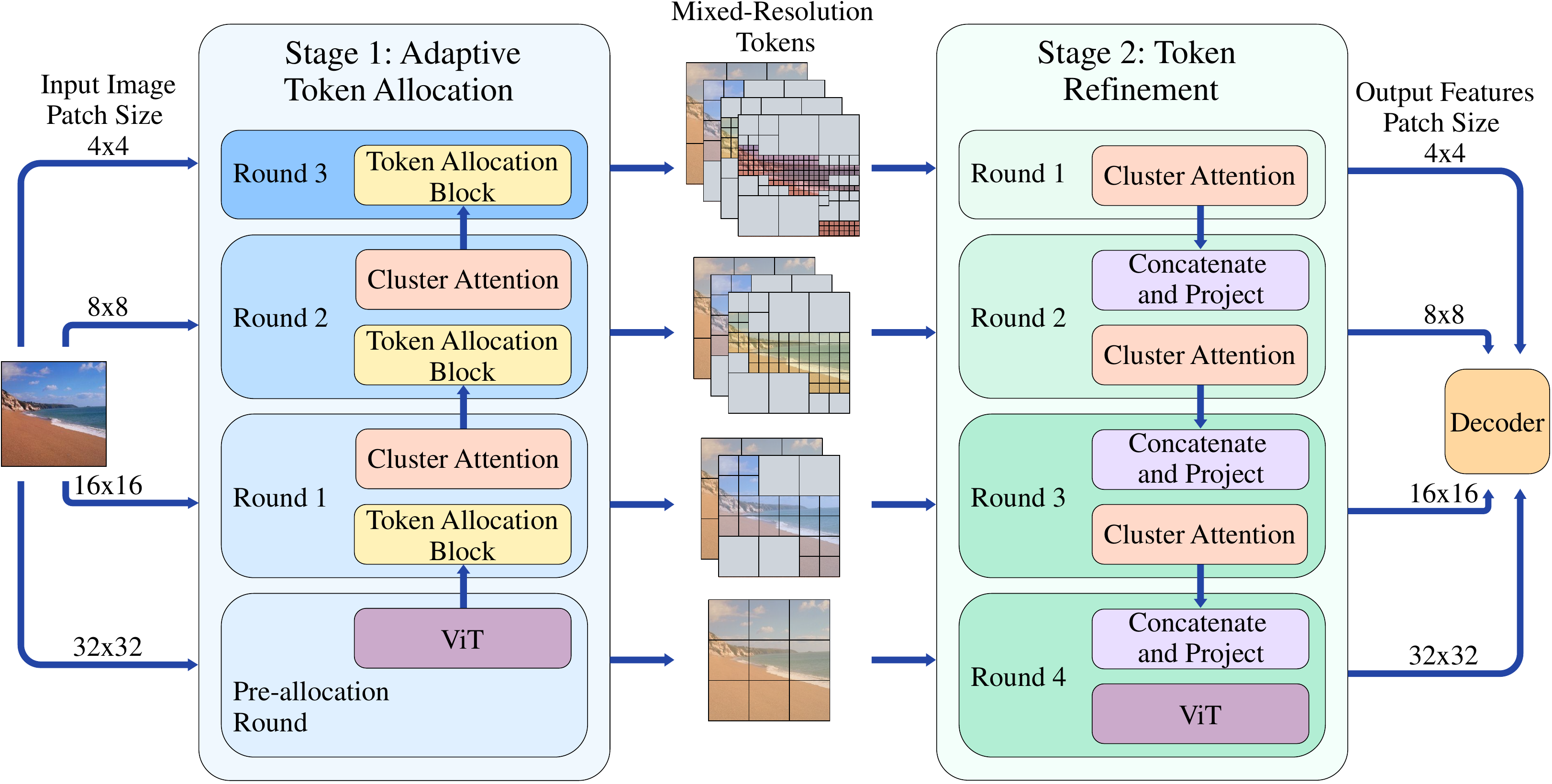}
  \caption{ARTA overview. ARTA has two stages: Adaptive Token Allocation (Stage~1) and Token Refinement (Stage~2). Stage~1 proceeds bottom-up from coarse $32{\times}32$ tokens. A pre-allocation ViT produces boundary-aware features, after which each allocation round applies a token allocation block (rounds 1--2 also use cluster attention). The allocation block scores the finest tokens and allocates finer-grained tokens to the corresponding patches containing class-boundaries (Figure~\ref{fig:dyn_up_block}), progressively building mixed-resolution sets up to $[32{\times}32,16{\times}16,8{\times}8,4{\times}4]$. Stage~2 proceeds top-down from the final set and refines tokens using cluster attention, with a ViT in the final round. At each refinement round, the finest tokens are output, and the remaining tokens continue. Lateral Stage~1 outputs are fused into Stage~2 by concatenation and projection before attention. The decoder uses these multi-scale features for dense prediction.}
  \label{fig:arta_architecture}
  \end{center}
\end{figure}

ARTA follows a two-stage design. In Stage 1, an adaptive mixed-resolution token allocator iteratively predicts a class-boundary score for tokens and hierarchically allocates additional high-resolution tokens to the image patches containing class-boundaries over multiple allocation rounds, producing a mixed-resolution token set. In Stage 2, a deeper hierarchical encoder refines token features over multiple refinement rounds to capture higher-level semantics using mixed-resolution attention, allowing coarse and fine tokens to interact while concentrating compute on semantically rich patches. The overall architecture is illustrated in Figure~\ref{fig:arta_architecture} and configurations details can be found in Table~\ref{tab:arta_settings}.


\begin{figure}[t]
  \begin{center}
  \includegraphics[width=0.97\textwidth]{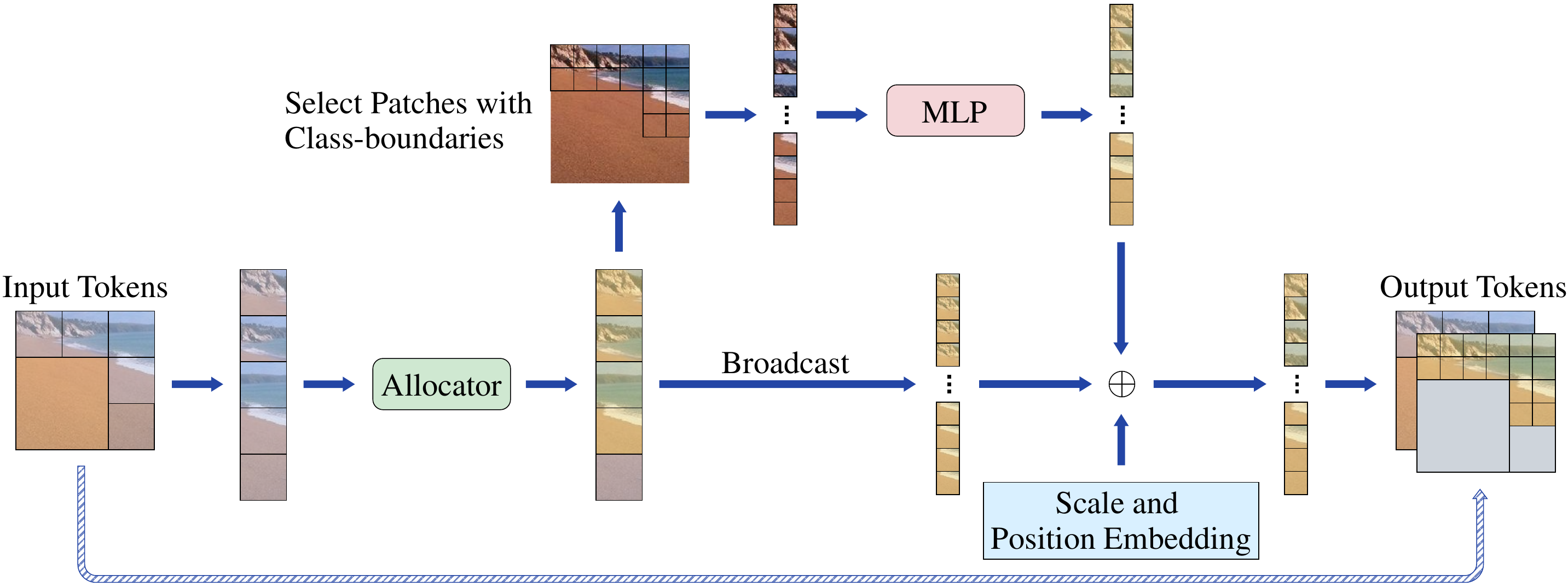}
  \caption{Token Allocation Block. The allocator scores the current finest-resolution tokens and selects the corresponding patches containing class-boundaries. Selected patches are split into $2{\times}2$ sub-patches to allocate finer tokens. Each new token is initialized from sub-patch image features (MLP) and combined with a broadcast residual from the parent token. Finally, scale and position embeddings are added to get output tokens.}
  \label{fig:dyn_up_block}
  \end{center}
\end{figure}

\subsection{Adaptive Mixed-Resolution Token Allocation}
\label{sec:arta_allocation}

\subsubsection{Coarse initialization.}
Given an input image, ARTA begins with a coarse patch embedding using a $32\times32$ kernel to form an initial low-resolution token grid. During a pre-allocation round, these tokens are processed by a lightweight Vision Transformer to produce boundary-aware coarse features, enabling reliable class-boundary scoring during the allocation rounds.

\subsubsection{Hierarchical allocation rounds.}
Stage 1 allocates tokens through $R$ allocation rounds (we use $R=3$). At allocation round $r$, the allocator predicts a class-boundary score $c_i^{(r)}$ for the candidate patch represented by token $i$ at the current finest resolution under consideration. Patches with scores above a low threshold are selected for finer allocation, and additional tokens are introduced by partitioning each selected region into finer sub-patches (in our implementation, a $2\times2$ partition yielding four tokens per selected patch). 

The resulting mixed-resolution token set is then processed by a cluster attention block from AFF~\cite{ziwen_autofocusformer_2023}, which computes attention over localized clusters that are recomputed each forward pass under the current allocation.
While AFF adaptively downsamples by merging less important tokens into important ones, yielding non-overlapping tokens, ARTA can contain overlapping tokens at multiple resolutions covering the same region. We use the same clustering mechanism to enable efficient cross-resolution interaction, so that fine tokens can attend to nearby coarse tokens for high-level context, while coarse tokens absorb local detail carried by fine tokens that inject newly extracted sub-patch features.

At the next round, the allocator is re-applied only to the finest-resolution tokens produced by the previous allocation round, rather than re-scoring the entire token set, reducing allocation overhead by avoiding repeated dense evaluation.

\subsubsection{Mixed-resolution token set.}
After $R$ rounds, Stage 1 produces a mixed-resolution representation consisting of coarse tokens for homogeneous regions and finer tokens concentrated around predicted semantic boundaries. This ensures that no region is represented at a higher resolution than necessary.

\subsection{Token Allocation Block}
\label{sec:token_allocation_block}

The token allocation block assigns a scalar class-boundary score to each token at the finest resolution in allocation round $r$. For an input token $i$, the allocator outputs a score $c_i^{(r)}$ estimating the amount of semantic (class) boundary content within the token patch.

\subsubsection{Class-boundary score.}
\label{sec:class_boundary_score}

The allocator is trained to predict a class-boundary score for each token, indicating how important it is to allocate additional tokens to the corresponding image patch. Ground-truth scores are derived from the segmentation labels by first computing a binary map of class-boundary pixels, where a pixel is marked as a boundary (1) if any of its neighboring pixels has a different class label, and as non-boundary (0) otherwise. For each token patch, we sum the boundary pixels within the patch and normalize the result to obtain the target score. The allocator MLP is trained to regress this target using mean squared error (MSE).

This scoring choice encourages each token to represent a single semantic class, simplifying downstream reasoning. A token that spans multiple classes must encode both class identities and their spatial arrangement, increasing representational burden. In contrast, single-class tokens allow the encoder to focus on semantic meaning rather than intra-token class composition.

\subsubsection{Threshold-based allocation.}
\label{sec:threshold_allocation}

We allocate additional tokens only for regions whose predicted scores exceed a round-specific threshold $\tau_r$. This induces a variable token budget per sample. Let $N_r$ denote the number of candidate tokens for a given sample at round $r$, and let
\begin{equation}
K_r = \sum_{i=1}^{N_r} \mathbf{1}\!\left(c_i^{(r)} > \tau_r\right)
\end{equation}
denote the number of selected tokens. The threshold $\tau_r$ is set slightly above zero to tolerate small prediction noise and to ensure that only regions with sufficient predicted boundary complexity are selected for further allocation.

Because $K_r$ varies across samples, token counts differ within a mini-batch. During training, we pad each sample's token set (per allocation round) to the maximum token count in the batch and mask the padded tokens so they do not participate in attention or loss computation. This enables mini-batch training while ensuring each sample uses only the number of tokens required by its content.

\subsubsection{Token allocation.}
\label{sec:token_allocation}

For each selected token, the corresponding image patch is partitioned into $2\times2$ sub-patches, and allocate four finer-grained tokens, one per sub-patch. To initialize a new token, we extract the raw pixel values of its sub-patch, flatten them into a vector, and project this vector to the token embedding dimension using a learned linear layer. The projected vector is then passed through an MLP to produce an image feature. To reuse the selected token’s boundary-aware coarse representation, we add it as a residual to each new token. Each token further receives (i) a learned scale embedding and (ii) a learned sub-patch position embedding to disambiguate within-patch spatial identity.

\begin{table}[t]
\caption{Encoder architecture hyperparameters for ARTA-Tiny, ARTA-Small, and ARTA-Base. For each round, we report the token embedding dimensions and the number of attention blocks. Entries are listed in execution order, following ARTA's coarse-to-fine-to-coarse layout.}
\label{tab:arta_settings}
\begin{center}
\small
\setlength{\tabcolsep}{6pt}
\begin{tabular}{l|ll|ll}
\hline
\multicolumn{1}{c|}{}
&\multicolumn{2}{c|}{Stage 1}
&\multicolumn{2}{c}{Stage 2} \\
\multicolumn{1}{c|}{Encoder}
&\multicolumn{2}{c|}{Patch Size: [$32^2$, $16^2$, $8^2$, $4^2$]}
&\multicolumn{2}{c}{Patch Size: [$4^2$, $8^2$, $16^2$, $32^2$]} \\
\multicolumn{1}{c|}{ } 
& \multicolumn{1}{c}{Dims} 
& \multicolumn{1}{c|}{Blocks} 
& \multicolumn{1}{c}{Dims} 
& \multicolumn{1}{c}{Blocks} \\
\hline
\hline
ARTA-Tiny  & [512, 256, 128, 64] & [1, 1, 1, 0] & [64, 128, 256, 512] & [4, 4, 16, 4]\\  
ARTA-Small & [512, 256, 128, 64] & [2, 2, 2, 0] & [64, 128, 256, 512] & [4, 6, 24, 3]\\  
ARTA-Base  & [768, 384, 192, 96] & [2, 2, 2, 0] & [96, 192, 384, 768] & [8, 6, 18, 4]\\ 
\hline
\end{tabular}
\end{center}
\end{table}

\subsection{Mixed-Resolution Token Refinement}
\label{sec:arta_refinement}

Stage~2 takes the mixed-resolution token set produced by the final allocation round in Stage~1 and refines token features in a top-down manner to capture higher-level semantics. Stage~2 is organized into multiple refinement rounds. In each round, we apply cluster attention~\cite{ziwen_autofocusformer_2023} over localized token clusters, which is computationally efficient for sparse mixed-resolution token sets and enables cross-resolution interactions. Fine tokens gain broader semantic context by attending to nearby coarse tokens, while coarse tokens are enriched with local structure information carried by fine tokens.

Across refinement rounds, Stage~2 progressively outputs the finest-resolution tokens to form multi-scale features for dense prediction. Starting from mixed-resolution tokens at all four scales, each refinement round outputs the current finest scale (e.g., $4{\times}4$, then $8{\times}8$, then $16{\times}16$, then $32{\times}32$), while the remaining coarser tokens are passed to the next round. The final refinement round uses a ViT block to strengthen high-level semantic representations at the coarsest scale.

\subsubsection{Cross-stage residual connections.}
To preserve low-level and boundary-aware information from the allocation stage, Stage~2 incorporates lateral connections from Stage~1. Before the attention block in each refinement round, we fuse the current Stage~2 tokens with a Stage~1 lateral output at the same resolution via concatenation followed by linear projection. Concretely, refinement rounds 2--4 use lateral outputs from allocation round~2, allocation round~1, and the pre-allocation round, respectively. This pre-attention fusion injects Stage~1 features into Stage~2 refinement while maintaining the mixed-resolution structure. The output finest-scale tokens from each refinement round are forwarded to the decoder as multi-scale features.

\label{sec:decoder_strategy}
\subsection{Decoder strategy}

We use the point-based Mask2Former decoder~\cite{cheng_masked-attention_2022} as in AFF~\cite{ziwen_autofocusformer_2023}. A pixel decoder first processes and aligns features across scales, followed by masked cross-attention that iteratively refines a set of class queries using multi-scale token features.

Before masked cross-attention, we densify only the highest-resolution feature map. At each spatial location, we select the finest token available, replicate its feature over the corresponding region, and add learned positional embeddings for spatial differentiation. Lower-resolution feature maps remain sparse. This preserves sparse computation in early decoding while enabling dense, fine-grained updates in the final masked attention stage.

\section{Experiments}

\subsection{Datasets}
We evaluate our method on three publicly available semantic segmentation benchmarks: ADE20K \cite{zhou_semantic_2019}, Cityscapes \cite{cordts_cityscapes_2016}, and COCO-Stuff \cite{lin_microsoft_2014, caesar_coco-stuff_2018}.
ADE20K is a scene parsing dataset containing 20,210 images annotated with 150 fine-grained semantic classes, covering a diverse range of indoor and outdoor environments.
Cityscapes is a street-level driving dataset consisting of 5,000 high-resolution images with fine annotations for 19 semantic categories.
COCO-Stuff extends the COCO dataset with dense pixel-level annotations, providing 172 semantic labels across 164,000 images.

\begin{table}[t]
\caption{Comparisons with state-of-the-art methods on ADE20K val. }
\label{tab:ade20k}
\begin{center}
\begin{tabular}{l|c|c|cc}
\hline
\multirow{2}{*}{Model}
&\multirow{2}{*}{Params}
&\multirow{2}{*}{FLOPs $\downarrow$}
&\multicolumn{1}{c}{mIoU }
&\multicolumn{1}{c}{mIoU }
\\ 
 &  &  & \multicolumn{1}{c}{(SS) $\uparrow$} & \multicolumn{1}{c}{(MS) $\uparrow$} \\
\hline
\hline
ViT-CoMer-T \cite{xia_vit-comer_2024} & 38.7M & - & 43.0 & 44.3 \\
SegFormer-B3 \cite{xie_segformer_2021} & 47.3M & 79G & 49.4 & 50.0 \\
SegNeXt-L \cite{guo_segnext_2022} & 48.9M & 70G & 51.0 & 52.1 \\
SegMAN-B \cite{fu_segman_2025} & 51.8M & 58G & \textbf{52.6} & - \\
Mask2Former-Swin-T \cite{cheng_masked-attention_2022} & 46.5M & 74G & 47.7 & 49.6 \\
AFF-Tiny-1/5 \cite{ziwen_autofocusformer_2023} & 46.5M & 51G & 50.0 & -  \\
ARTA-Tiny & 48.5M & \textbf{44$\pm$7G} & 51.5 & \textbf{52.6} \\
\hline
ViT-CoMer-S \cite{xia_vit-comer_2024} & 61.4M & - & 46.5 & 47.7 \\
VMamba-T \cite{liu_vmamba_2024} & 62.0M & - & 47.9 & 48.8 \\
ViT-Adapter-S \cite{chen_vision_2022} & 57.6M & - & 46.2 & 47.1 \\
OverLoCK-Tiny \cite{lou_overlock_2025} & 63.0M & - & 50.3 & - \\
HRFormer-B \cite{yuan_hrformer_2021} & 56.2M & - & 48.7 & 50.0 \\
SegFormer-B4 \cite{xie_segformer_2021} & 64.1M & 96G & 50.3 & 51.1 \\
LRFormer-B \cite{wu_low-resolution_2025} & 69M & 75G & 51.0 & - \\
Mask2Former-Swin-S \cite{cheng_masked-attention_2022} & 66.5M & 98G & 51.3 & 52.4 \\
AFF-Small-1/5 \cite{ziwen_autofocusformer_2023} & 62.1M & 67G & 51.9 & - \\
ARTA-Small & 61.7M & \textbf{60.7$\pm$9.3G} & \textbf{52.1} & \textbf{53.9} \\
\hline
ViT-CoMer-B \cite{xia_vit-comer_2024} & 144.7M & - & 48.8 & 49.4 \\
ViT-Adapter-B \cite{chen_vision_2022} & 133.9M & - & 48.8 & 49.7 \\
VMamba-B \cite{liu_vmamba_2024} & 122.0M & - & 51.0 & 51.6 \\
ConvNeXt V2-B \cite{woo_convnext_2023} & 122.0M & - & 52.1 & - \\
OverLoCK-Base \cite{lou_overlock_2025} & 124M & - & 51.7 & - \\
SegFormer-B5 \cite{xie_segformer_2021} & 84.7M & 183G & 51.0 & 51.8 \\ 
LRFormer-L \cite{wu_low-resolution_2025} & 113M & 183G & 52.6 & - \\
SegMAN-L \cite{fu_segman_2025} & 92.4M & 97G & 53.2 & - \\
Mask2Former-Swin-B \cite{cheng_masked-attention_2022} &  106.5M & 222.7G & 52.4 & 53.7 \\
ARTA-Base & 111.5M & \textbf{82.4$\pm$14.0G} & \textbf{53.5} & \textbf{54.6}\\
\hline
\end{tabular}
\end{center}
\end{table}

\subsection{Experimental Setup}
\label{sec:experimental_setup}
We build on Detectron2 with components from AFF and Mask2Former, and train on NVIDIA A100 GPUs. For complete details of the experimental protocol and additional settings, see the supplementary material.

For pre-training, we use ImageNet classification. During this stage the token allocation blocks are disabled and tokens are allocated at random according to a fixed ratio schedule (no content-based selection). ImageNet classification results can be found in the supplement.

For fine-tuning, we follow AutoFocusFormer and Mask2Former for data augmentation and general hyperparameters. Models were optimized using AdamW with a learning rate of $4\times10^{-5}$. Crop sizes are fixed for all reported methods: 512$\times$512 for ADE20K and COCO-Stuff, and 1024$\times$1024 for Cityscapes. Training was conducted for 80k iterations with a batch-size of 32 on ADE20K and COCO-Stuff, and for 90k iterations on Cityscapes with a batch-size of 16. The threshold $\tau_r$ is set to [0.005, 0.01, 0.02] for the three allocation rounds.

For evaluation on ADE20K and COCO-Stuff, we resize the short side to 512 with preserved aspect ratio; for Cityscapes, we use overlapping 1024$\times$1024 sliding-window inference. Performance is reported using mean Intersection over Union (mIoU)

FLOPs are measured for the full network at fixed input sizes: 512$\times$512 (ADE20K/COCO-Stuff) and 1024$\times$2048 (Cityscapes), reported as mean~$\pm$~std per image over the validation set. We compute FLOPs for \textbf{ARTA} using this protocol, while FLOPs for other methods are taken from their respective papers. For fairness, we report FLOPs only for methods evaluated at the same input resolution.

\begin{table}[ht]
\caption{Comparisons with state-of-the-art methods on the validation sets of COCO-Stuff and Cityscapes. ``$\dagger$'' indicates that ImageNet22k was used for pre-training.}
\label{tab:coco_cityscapes}
\begin{center}
\small
\setlength{\tabcolsep}{3pt}
\begin{tabular}{l|c|cl|cl}
\hline
\multirow{3}{*}{Model}  
&\multirow{3}{*}{Params}
&\multicolumn{2}{c|}{COCO-Stuff}
&\multicolumn{2}{c}{Cityscapes} \\
\multicolumn{1}{c|}{}
& \multicolumn{1}{c|}{}
&\multicolumn{1}{c}{FLOPs $\downarrow$}
&\multicolumn{1}{c|}{mIoU $\uparrow$}
&\multicolumn{1}{c}{FLOPs $\downarrow$}
&\multicolumn{1}{c}{mIoU $\uparrow$}\\
\multicolumn{1}{c|}{}
& \multicolumn{1}{c|}{}
&\multicolumn{1}{c}{}
&\multicolumn{1}{c|}{(SS/MS)}
&\multicolumn{1}{c}{}
&\multicolumn{1}{c}{(SS/MS)}\\\hline\hline
SegFormer-B3 \cite{xie_segformer_2021} & 47.3M & 79G & 45.5 / - & 963G & 81.7 / 83.3 \\
SegNext-L \cite{guo_segnext_2022} & 48.9M & 70G & 46.5 / 47.2 & 578G & 83.2 / \textbf{83.9}  \\
SegMAN-B \cite{fu_segman_2025} & 51.8M & 58G & \textbf{48.4}/ - & 479G & \textbf{83.8} / - \\
Mask2Former-Swin-T \cite{cheng_masked-attention_2022} & 46.5M & - & \hspace{4mm} - / - & 537G & 82.1 / 83.0 \\
ARTA-Tiny & 49.5M & \textbf{45$\pm$9G} & 47.2 / 47.7 & \textbf{286$\pm$20G} & 81.8 / 82.9 \\
\hline 
HRFormer-B \cite{yuan_hrformer_2021} & 56.2M & 280G & 42.4 / 43.3 & 2224G & 81.9 / 82.6  \\
SegFormer-B4 \cite{xie_segformer_2021} & 64.1M & 96G & 46.5 / - & 1241G & 82.3 / \textbf{83.9}  \\
LRFormer-B \cite{wu_low-resolution_2025} & 67M & 75G & 47.2 / - & 555G & \textbf{83.0} / -  \\
Mask2Former-Swin-S \cite{cheng_masked-attention_2022} & 66.5M & - & \hspace{4mm} - / - & 732G & 82.6 / 83.6 \\
ARTA-Small & 61.7M & \textbf{57$\pm$12G} & \textbf{47.6} / \textbf{48.3} & \textbf{355$\pm$25G} & 81.9 / 83.2 \\
\hline 
SegFormer-B5 \cite{xie_segformer_2021} & 84.7M & 112G & 46.7 / - & 1460G & 82.4 / 84.0  \\ 
LRFormer-L \cite{wu_low-resolution_2025} & 111.0M & 122G & 47.9 / - & 908G & 83.2 / - \\
SegMAN-L \cite{fu_segman_2025} & 92.4M & 97G & 48.8 / - & 796G & \textbf{84.2} / - \\
Mask2Former-Swin-B$\dagger$ \cite{cheng_masked-attention_2022} &  106.5M & - & \hspace{4mm} - / - & 1050G & 83.3 / \textbf{84.5} \\
ARTA-Base & 111.5M & \textbf{87$\pm$19G} & \textbf{49.0} / \textbf{49.4} & \textbf{590$\pm$40G} & 82.6 / 83.3\\
\hline
\end{tabular}
\end{center}
\end{table}

\subsubsection{ADE20K.}
We compare \textbf{ARTA} at three model scales against state-of-the-art baselines on ADE20K val (Table~\ref{tab:ade20k}). At the smallest scale, SegMAN-B attains higher single-scale mIoU than ARTA-Tiny, but ARTA-Tiny is more compute-efficient. As model capacity increases, ARTA scales more favorably and at the large scale ARTA-Base surpasses SegMAN-L while using fewer FLOPs. When comparing to other encoders using a Mask2Former decoder such as Mask2Former~\cite{cheng_masked-attention_2022} with Swin backbones and AFF~\cite{ziwen_autofocusformer_2023}, we can observe that ARTA has higher mIoU and lower FLOPs over all model sizes. Across all ARTA variants, multi-scale testing further boosts performance, with ARTA-Base reaching \textbf{54.6} mIoU.

\subsubsection{COCO-Stuff.}
Table~\ref{tab:coco_cityscapes} mirrors the trend observed on ADE20K: SegMAN is strongest at the smallest scale, whereas ARTA improves more with capacity. In the $\sim$50M-parameter regime, SegMAN-B achieves higher single-scale mIoU than ARTA-Tiny, but ARTA-Tiny is notably more compute-efficient. As model size increases, ARTA gains substantially more mIoU from Tiny to Base than SegMAN does from B to L. At the large scale, \textbf{ARTA-Base} surpasses SegMAN-L in single-scale mIoU while using fewer FLOPs on average, and further improves with multi-scale testing (Table~\ref{tab:coco_cityscapes}).

\subsubsection{Cityscapes.}
On Cityscapes, ARTA delivers competitive accuracy while using a fraction of the compute of comparable baselines (Table~\ref{tab:coco_cityscapes}). Across model scales, ARTA attains mIoU close to state of the art, but with substantially lower FLOPs (e.g., \textbf{ARTA-Tiny} at 286\,GFLOPs vs.\ 479–963\,GFLOPs for similar-size baselines), providing a favorable accuracy–efficiency trade-off.

\subsubsection{Speed Analysis.}

We benchmark inference speed and memory usage on Cityscapes with a single NVIDIA A100 GPU. Following SegMAN~\cite{fu_segman_2025}, we report FPS averaged over 128 inference steps with batch size 2, using full-resolution inputs (1024$\times$2048) in FP32. We report peak allocated GPU memory over the same run using PyTorch’s CUDA memory statistics. As shown in Table~\ref{tab:fps_mem}, ARTA’s reduced FLOPs translate to higher throughput and lower memory usage across model scales.

\begin{figure}[t]
  \begin{center}
  \includegraphics[width=0.99\textwidth]{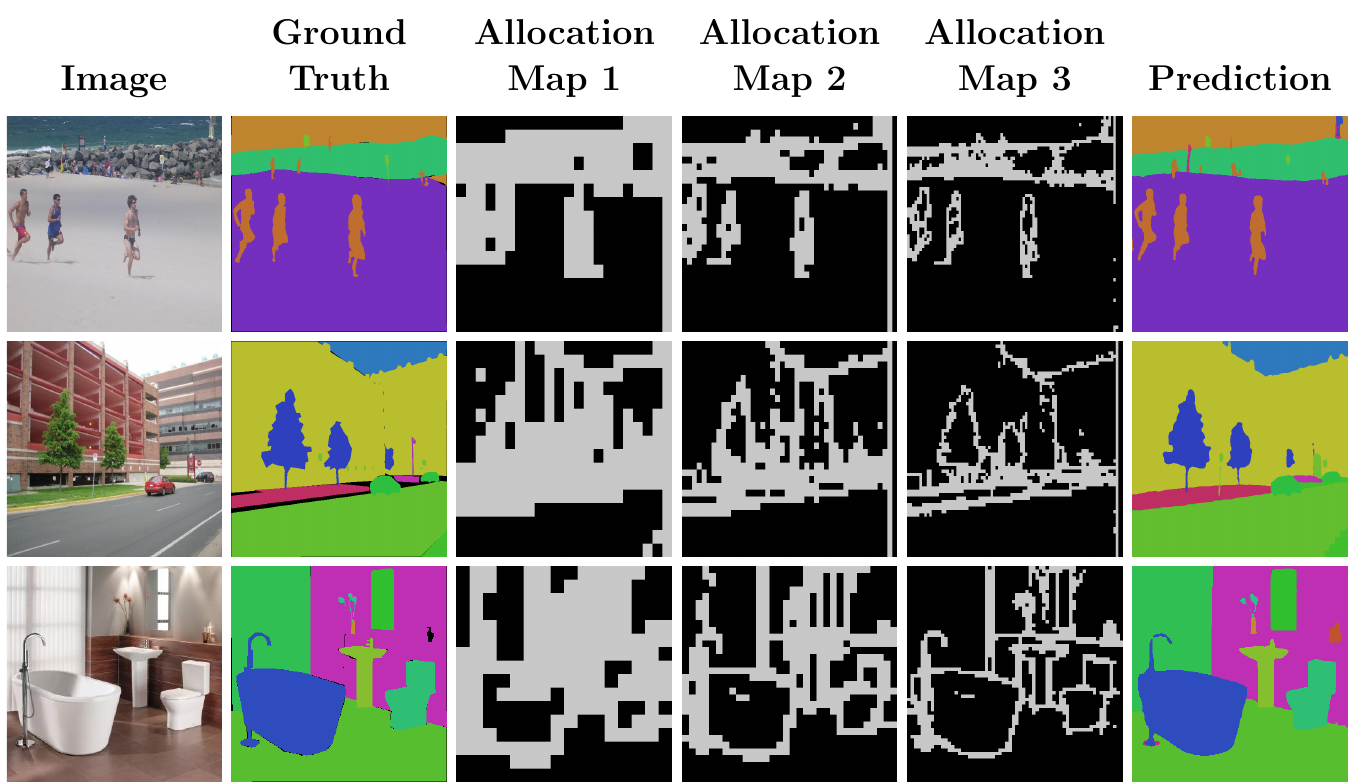}
  \caption{From left to right: original image, ground truth, $32 \times 32$ patches selected for allocation, $16 \times 16$ patches selected for allocation, $8 \times 8$ patches selected for allocation, and prediction. Black in the ground truth means that the pixel was not labeled.}
  \label{fig:qual_ex}
  \end{center}
\end{figure}

\subsubsection{Qualitative Examples.}

Figure~\ref{fig:qual_ex} shows qualitative results and the tokens selected for refinement at each scale. The model keeps homogeneous regions (e.g., sky, road, walls) coarse while allocating finer tokens near boundaries and small/fine structures, aligning with the goal of spending resolution where semantic complexity is high.

\begin{table}[t]
\centering
\setlength{\tabcolsep}{5pt}
\caption{FPS and peak allocated memory during inference on Cityscapes with a  single NVIDIA A100.}
\label{tab:fps_mem}
\begin{tabular}{lcccc}
\hline
\multicolumn{1}{c}{Model}
&\multicolumn{1}{c}{Params}
&\multicolumn{1}{c}{FLOPs} 
&\multicolumn{1}{c}{FPS}
&\multicolumn{1}{c}{Peak mem (GB)}\\
\hline
\hline
SegFormer-B3 \cite{xie_segformer_2021} & 47.3M & 963G & 6.6 & 21.9  \\
SegNeXt-L \cite{guo_segnext_2022} & 48.8M & 554G & \textbf{7.3} & 16.9  \\
SegMAN-B \cite{fu_segman_2025} & 51.8M & 479G & 5.6 & 23.8  \\
AFF-Tiny \cite{ziwen_autofocusformer_2023} & 46.5M & 463G & 3.8 & 34.4  \\
ARTA-Tiny & 49.5M & \textbf{286$\pm$20G} & 6.6 & \textbf{14.6}  \\
\hline
SegFormer-B4 \cite{xie_segformer_2021} & 64.1M & 1241G & 4.7 & 29.5  \\
AFF-Small \cite{ziwen_autofocusformer_2023} & 62.1M & 639G & 3.6 & 38.7  \\
ARTA-Small & 61.7M & \textbf{355$\pm$25G} & \textbf{6.0} & \textbf{19.8}  \\
\hline
SegFormer-B5 \cite{xie_segformer_2021} & 84.7M & 1460G & 4.0 & 34.9  \\
SegMAN-L \cite{fu_segman_2025}  & 92.4M & 796G & 3.9 & 31.6  \\
ARTA-Base & 111.5M & \textbf{590$\pm$40G} & \textbf{4.9} & \textbf{25.4}  \\
\hline
\end{tabular}
\end{table}

\subsection{Ablation Studies}

\subsubsection{Without adaptive token allocation}

We ablate adaptive token allocation by forcing dense tokenization: during ImageNet pre-training, every selected coarse token is always allocated finer tokens, producing dense token grids at all scales. We then fine-tune this ARTA-Tiny variant on ADE20K and compare it to the adaptive baseline under the same training setup. Removing adaptive allocation reduces accuracy and increases compute, achieving 50.4 mIoU at 74G FLOPs versus 51.5 mIoU at 44$\pm$7G for the baseline.

\subsubsection{Choice of encoder blocks.}

We ablate the block type used at different locations in ARTA-Tiny, trained from scratch on ADE20K. The pre-allocation and final refinement rounds operate on coarse and dense token grids, where standard dense backbones are applicable. In contrast, all other rounds process sparse mixed-resolution token sets with substantially higher token counts, requiring layers that can handle multi-scale sparse inputs while keeping compute low. A vanilla ViT supports sparse inputs in principle, but becomes infeasible in the intermediate rounds due to its quadratic attention cost. We therefore consider two compatible alternatives that support sparse mixed-resolution processing with compute constraints: cluster attention and (point-based) multi-scale deformable attention (MSDeformAttn)~\cite{ziwen_autofocusformer_2023}. Table~\ref{tab:ablation_attention} shows that ViT is best for the coarse dense initial/final rounds, while cluster attention performs best for intermediate mixed-resolution rounds.

\begin{table}[t]
\centering
\setlength{\tabcolsep}{6pt}
\small
\caption{Block-type ablation for ARTA-Tiny on ADE20K. ``OOM'' indicates out of memory.}
\label{tab:ablation_attention}
\begin{tabular}{l|l|c}
\hline
Initial/Final block type &  Intermediate block type & \multirow{1}{*}{mIoU} \\
\hline \hline
\multicolumn{3}{c}{Initial/Final Block Ablation} \\
\hline
Cluster Attention~\cite{ziwen_autofocusformer_2023} & \multirow{4}{*}{Cluster Attention~\cite{ziwen_autofocusformer_2023}} & 42.1 \\
ConvNeXt V2~\cite{woo_convnext_2023}           &  & 41.8 \\
Swin~\cite{liu_swin_2021}                      &  & 42.2 \\
ViT~\cite{dosovitskiy_image_2020}              &  & \textbf{42.4} \\
\hline
\multicolumn{3}{c}{Intermediate Block Ablation} \\
\hline
\multirow{3}{*}{ViT~\cite{dosovitskiy_image_2020}} & ViT~\cite{dosovitskiy_image_2020} & OOM \\
& MSDeformAttn~\cite{ziwen_autofocusformer_2023} & 42.0 \\
& Cluster Attention~\cite{ziwen_autofocusformer_2023} & \textbf{42.4} \\
\hline
\end{tabular}
\end{table}

\subsubsection{Ablation: Oracle Token Allocation Scores. }
We test whether injecting ground-truth class-boundary scores during training improves segmentation. Models are fine-tuned on ADE20K; the allocation blocks are always trained and, at inference, allocation uses predicted scores. An oracle rate of $x\%$ means we randomly use oracle scores for $x\%$ of training batches (and predicted scores otherwise). Results on validation set are shown in Table~\ref{tab:oracle}.

Oracle scores do not improve validation mIoU and higher oracle rates degrade performance. While training losses (Dice/mask) decrease when using oracle scores, we hypothesize this reflects a train--test mismatch: oracle scores may act as a privileged boundary cue during training, encouraging the model to rely on boundary information that is not available at test time. When allocation reverts to predicted scores at inference, this reliance may not transfer and performance drops.

\begin{table}[t]
\centering
\small
\begin{minipage}[t]{0.48\textwidth}\centering
\caption{Oracle token allocation scores on ADE20K with ARTA-Tiny.}
\label{tab:oracle}
\begin{tabular}{cc}
\hline 
\multicolumn{1}{c}{Oracle rate}
&\multicolumn{1}{c}{mIoU $\uparrow$} \\
\hline 
\hline 
100\% & 35.3 \\
50\% & 48.9 \\
10\% & 50.8 \\
0\% & \textbf{51.5} \\
\hline
\end{tabular}
\end{minipage}\hfill
\begin{minipage}[t]{0.48\textwidth}\centering
\centering
\caption{Ablation of Stage~2 and Stage~1 token initialization for ARTA-Tiny on ADE20K.}
\label{tab:ablation_network}
\begin{tabular}{lc}
\hline 
\multicolumn{1}{c}{Method }
&\multicolumn{1}{c}{mIoU $\uparrow$}
\\ \hline 
\hline 
Baseline (Stage 1 + Stage 2) & \textbf{42.4} \\
Stage 1 only & 41.6 \\
w/o aux image init & 41.0 \\
w/o feature residual & 41.3 \\
\hline
\end{tabular}
\end{minipage}
\end{table}

\subsubsection{Ablation: Stage design and token initialization.}
We ablate Stage~2 mixed-resolution token refinement and two token initialization choices in Stage~1 allocation, training all variants from scratch on ADE20K for 80k iterations. The baseline uses Stage~1 allocation followed by Stage~2 refinement. \textit{Stage~1 only} removes Stage~2, reallocates depth to Stage~1 to match parameters, and adds a final cluster-attention layer (as in Stage~2) to refine newly allocated high-resolution tokens. \textit{No aux image data} disables adding sub-patch image features when initializing newly allocated tokens. \textit{No feature residual} removes the residual copy of the selected token feature, initializing new tokens solely from sub-patch image features (plus embeddings/MLP). As shown in Table~\ref{tab:ablation_network}, removing Stage~2 or either initialization component reduces mIoU.

\section{Conclusion}

We presented ARTA, a coarse-to-fine segmentation backbone that predicts semantic boundary density early and allocates additional tokens to regions needing higher spatial detail. Starting from coarse tokens and allocating finer tokens adaptively, ARTA preserves compute by avoiding fine processing in homogeneous areas while maintaining interacting mixed-resolution features across scales. ARTA achieves state-of-the-art accuracy on ADE20K and COCO-Stuff with substantially fewer FLOPs, and remains competitive on Cityscapes at a fraction of the compute of comparable baselines. These results show that content-adaptive resolution is a strong approach for accurate, efficient dense prediction. Scalability to much larger backbones and longer pre-training remains untested. Future work includes extending ARTA to 3D segmentation.




%
%
\bibliographystyle{splncs04}
\bibliography{main}
\end{document}